\documentstyle[12pt]{article}

\newcounter{cex}
\newcounter{cexi}
\newcounter{cexii}

\def\ex{\refstepcounter{cex} \setcounter{cexi}{0} \item}
\def\exi{\refstepcounter{cexi} \setcounter{cexii}{0} \item}

\newenvironment{el}
{
 \begin{list}{(\arabic{cex})}{
    \topsep = 1.2ex
    \itemsep = 1.0ex
    \parsep = 0pt
    \rightmargin = 1.0em
    \leftmargin = 3.7em
    \labelwidth = 2.5em
    \labelsep = 1.5em
 }}{ \end{list} }

\newenvironment{eli}
{
 \begin{list}{\alph{cexi}.}{
    \itemsep = 0pt
    \parsep = 0pt
    \rightmargin = 0em
    \leftmargin = 1.4em
    \labelwidth = 1.0em
    \labelsep = 1.0em
 }}{ \end{list} }

\newlength{\badstar}

\title{\bf An NLP Approach to a Specific Type of Texts: Car Accident Reports}
\author{
\begin{normalsize}
\begin{tabular}{c c}
Dominique Estival                      & Fran\c{c}oise Gayral\\
ISSCO, Universit\'{e} de Gen\`{e}ve    & LIPN, Universit\'{e} Paris-Nord\\
54 rte des Acacias, CH-1227 Gen\`{e}ve & Av. J.-B. Cl\'{e}ment, F-93430
Villetaneuse\ \\
$<$estival@divsun.unige.ch$>$          & $<$fg@lipn.univ-paris13.fr$>$\\
+41-22-705-71-16                       & +33-1-49-40-36-25
\end{tabular}
\end{normalsize}
}
\date{ }

\begin{document}
\maketitle

\section{Introduction}

The work reported here is the result of a study done within a larger
project on the ``Semantics of Natural Languages'' viewed from the
field of Artificial Intelligence and Computational Linguistics.  In
this project, we have chosen a corpus of insurance claim
reports.\footnote{
The Project Inter-PRC {\em S\'{e}mantiques des Langues Naturelles},
sponsored by the French Minist\`{e}re de la Recherche, involves a
number of research centers and university laboratories.  The texts
were provided by the French insurance company MAIF, after being made
anonymous.  They have been translated into English by D.E.\ and the
original French texts are given in the annex.}

These texts deal with a relatively circumscribed domain, that of road
traffic, thereby limiting the extra-linguistic knowledge necessary to
understand them.  Moreover, these texts present a number of very
specific characteristics, insofar as they are written in a
quasi-institutional setting which imposes many constraints on their
production.

We first determine what these constraints are in order to then show
how they provide the writer with the means to create as succint a text
as possible, and in a symmetric way, how they provide the reader with
the means to interpret the text and to distinguish between its factual
and argumentative aspects.

\section{Characteristics of the texts}

This type of texts is culturally well-defined and possesses several
characteristics, which both the writer and the reader are perfectly
aware of when writing or reading one of them.  It is not a newspaper
story, nor a letter to a friend narrating the car accident, but an
insurance claim report which has to follow several constraints,
defined in \ref{text-param}.

\begin{el}
\ex Text Parameters
\begin{itemize}
\item[A.]  the text involves at least two participants, generally two
vehicles, one of which is the author's;

\item[B.]  the text is obligatorily short, at most one paragraph;

\item[C.] by definition, the text is a narration in which an accident
takes place;

\item[D.]  the text is sent to the author's insurance company.
\end{itemize}
\label{text-param}
\end{el}

Beyond these four parameters which are determined by the nature of the
reports, we also find in this text presuppositions due to the
particular domain involved, the ``road'' domain.  This domain-specific
knowledge, which is part of the more general context {\it C\/}, is
called here {\it K\/}.  {\it K\/} concerns vehicles, vehicle motions,
traffic rules, the usual behavior of drivers and pedestrians, and also
some elements of ``naive'' geometry.

Parameter D has a special bearing in so far as the writers know that
the insurance agents must pass a judgement on their behavior and will
determine their share of responsibility in the accident.  Necessarily,
the authors of those reports, while supposedly describing in an
impartial way the different events which have occurred, will attempt
to lessen their responsability.  The texts thus present many instances
of argumentative devices, whose usage forms part of the more general
knowledge of the language conventions, {\it LC\/}.  In a symmetric
way, the reader, i.e.\ the insurance agent, must untangle the factual
description from the argumentative presentation of the events.

We can define the tasks that this type of texts presents for the
writer and for the reader as in \ref{writer-prob} and
\ref{reader-prob} respectively.

\begin{el}
\ex The Writer's Problem:
The writer {\it W\/} knows the factual content {\it P\/} corresponding to
the circumstances of the accident and wants to convey it through a
text {\it T\/}.  {\it W\/} must then choose a {\it T\/} such that (a) it
will allow a reader {\it R\/} to rediscover {\it P\/}, and (b) it will
minimize {\it W\/}'s responsibility.
\label{writer-prob}
\end{el}

\begin{el}
\ex The Reader's Problem:
The reader {\it R\/} knows the language conventions {\it LC\/} and a part
of the context {\it C\/}.  {\it R\/} must then determine (a) the
factual content {\it P\/} of the text {\it T\/} and (b) the argumentation
presented by its writer {\it W\/}.
\label{reader-prob}
\end{el}

These two symmetrical tasks are thus both composed of a factual and an
argumentative part.  These two parts also coincide with the two goals
we can define for an NLP approach to understanding and processing
these texts.  At the first level, we try to extract from the text the
objective content corresponding to a factual analysis in order to
recreate the event:  ``What happened?  What real world events
concerning the motions of these vehicles or the scene geometry
actually occurred?''

At the second level, we take into account the nature and intent of the
text.  Our problem is then to uncover the argumentative devices used
by the writer and to determine how they can be used by the reader, and
later by our system, in interpreting the texts.

Accordingly, the remainder of this paper is divided into two parts.
In section \ref{Facts}, we show the importance of the situational and
cultural presuppositions for parameters A, B and C, and in section
\ref{Argument}, we take into account parameter D, which determines the
argumentative aspect of the texts.

\section{Factual content of the texts}
\label{Facts}

\subsection{Parameter A}
\label{param-A}

Parameter A (the fact that car accidents usually involve two
participants, most often two vehicles) is used to infer the identity
of some entities in the texts, or to establish coreference between two
entities.

There is a specific naming convention in French insurance claim
reports for the vehicles involved in an accident:  claimants must
refer to their own vehicle as ``A'' and to their opponent's as ``B''.
This convention, although it is part of the shared knowledge about
{\it LC\/}, is not always followed, and indeed it seems to be a burden
for {\it W\/}.  The reason is probably that a stereotypical
description using only labels ``A'' and ``B'' for the vehicles
involved sounds very neutral, such as could have been made by any
independant observer of the accident, while in fact {\it W\/} was
directly involved in the accident, and is thus personnally implicated, as
a person endowed with awareness and intentionality.  Thus, the authors
often do not seem to be able to choose between a narrative style using
the first person (``I'') and a descriptive style using the third
person (``vehicle A'').  Most of the texts are not homogeneous in this
respect and combine the two styles, as if there was a struggle,
probably unconscious, between a spontaneous narration of the different
events and a stereotyped description using the convention.

\begin{el}
\ex
{\it \underline{Vehicle A} waiting and stopped at the Pont de
Levallois lights.  Vehicle B arrived and hit  \underline{my} left side
mirror with its right side mirror.}
\label{A-B}
\end{el}

This hesitation which we observe in our texts is reinforced by the use
of metonymy.  Metonymy is often used to allow the identification of
the container with its content, and a common use of metonymy in our
texts concerns the vehicle and its driver.  Metonymy builds a unique
discourse entity, and as a consequence some properties from the
vehicle are transferred to the driver, and vice-versa.
Personalization of the vehicles is a typical example of this transfer:
a vehicle becomes endowed with the properties of a human being, for
instance by transferring from the driver to the car the property of
intentionality as in \ref{1}, or that of agentivity as in \ref{T7}.

\begin{el}
\ex
\begin{eli}
\exi {\it Vehicle B \underline{wanted} to turn right}
\label{1}

\exi {\it Being momentarily stopped in the right lane on Boulevard des
Italiens, I had switched my blinker on; I was at a stop and getting
ready to change lanes.  Vehicle B coming from my left
\underline{squeezed} too close to me and damaged the whole left
front side.}  (T7)
\label{T7}

\end{eli}
\end{el}

We can see from the example in \ref{right-door} that this unique
entity does not consist only of the vehicle and its driver, but can
also include its passengers.  In \ref{husband}, there is an
identification between the vehicle and the writer's husband, the
writer probably being the insured person.

\begin{el}
\ex
\begin{eli}
\exi {\it One of the cars in front of me opened its right front door} (T3)
\label{right-door}

\exi {\it My husband had entered the intersection when Mr.\ X's car hit
the front of the vehicle.}
\label{husband}
\end{eli}
\end{el}

Conversely, transference of properties can be made from the car to the
driver, as in \ref{bumper} or \ref{rolling}.  In \ref{bumper}, objects
(here the bumper) belonging to the vehicle are treated as belonging to
the driver.  In \ref{rolling}, the property of ``rolling along'' (the
literal meaning of the verb {\it rouler\/}) is transferred to the
driver.

\begin{el}
\ex
\begin{eli}
\exi {\it my bumper} (T11)
\label{bumper}

\exi {\it Je roulais} ({\it I was driving}, literally {\it I was rolling})
\label{rolling}
\end{eli}
\end{el}

This use of metonymy follows the coercion of semantic types (see
\cite{Pustejovsky89b}) in a predictable way:  the properties being
used to make an entity of one type (e.g.\ {\it car\/}:  ``inanimate
mechanical object'') into an entity of another type (e.g.\ {\it
driver\/}:  ``human agent'') are extractible in a regular way from the
predicate (e.g.\ {\it squeeze\/}:  ``requires an agentive subject'').

This coercion, economical for {\it W\/} because it allows a greater
conciseness, requires additional work on the part of {\it R\/}, because
{\it R\/} must make some inferences to undo it and to find the referent
of some expression.  For instance, a number of inferences, some of
them spatial, are necessary to find the identity of the agent, i.e.\ the
passenger in the right front seat, in \ref{right-door}.

Finding the exact referent of an expression is necessary in order to
block wrong inferences.  For instance, in \ref{T7}, the 1st person
refers successively to the car and to the driver:

\begin{itemize}
\item in {\it I had switched on my blinker\/}, the referent of {\it
I\/} is the driver;

\item {\it being stopped\/} can be understood with {\it I\/} referring
to the driver as well as to the vehicle;

\item in the last sentence, the word {\it me\/} must refer to the
vehicle:  from this text, {\it R\/} would never conclude that the
driver's left cheek had been bruised.
\end{itemize}

\subsection{Parameter B}

When setting to the task of writing such a report, the writer knows
parameter B, the constraint that only about a paragraph may be used to
relate the accident.\footnote{ There is a predefined area on the
printed form for the writer to write this text.}  At the same time,
{\it W\/} must not forget any important information whose absence
would prevent {\it R\/} from discovering the correct content {\it
P\/}.

So, {\it W\/} is thus faced with two goals:  to be exhaustive and to
be concise.\footnote{
Of course, this exhaustivity is not absolute, but is relative to the
context for these texts.  There is an ``informational norm'' in any
type of texts and the relevance of the information to be given is
judged relative to this informational norm, which is very difficult to
define precisely and which can only be known empirically (see
\cite{Kerbrat-Orecchioni86}).

The absence of information is as meaningful as its presence, but is
much harder to assess, since in order to bring it to light, we need a
basis for comparison.  In this context, the comparison could be
provided by the opponent's accident report; although difficult to
achieve, this would constitute an interesting study.}
These aims are not contradictory but force {\it W\/} to select the
information that will be given:  the text {\it T\/} must provide all the
information that is necessary in order to be understood, but only that
much.  We rediscover here Grice's Maxims \cite{Grice75}, in particular
the Maxim of Quantity, or Ducrot's exhaustivity law \cite{Ducrot72}.
Every detail mentioned by {\it W\/} can be assumed to be significant
and every adjective and adverb will carry some significance for the
narration, as in \ref{intense}, where extremely relevant modifiers are
piled up after the head noun.

\begin{el}
\ex
{\it the road on which the \underline{intense} traffic is going
\underline{one-way} in \underline{two lanes};} (T5)
\label{intense}
\end {el}

This constraint on the choice of information to give, which we call
``{\it W\/}'s selection problem'' and which we will later exploit to
infer some argumentative points, is part of the wider language
conventions {\it LC\/} and constitutes a ``meta-knowledge'', essential
for the success of communication.

Because of ``{\it W\/}'s selection problem'', {\it W\/} will generally
mention an event or an entity only in case its presence cannot be
deduced from {\it K\/} or from other types of shared background
knowledge, and only in case an explicit reference is absolutely
necessary to understand the text.  From this, it follows that the
number of entities introduced in the text will be kept to a minimum.
We can schematize this ``Minimality Assumption'' as follows:

\begin{el}
\ex Minimality Assumption:\\
  {\it W\/}'s selection problem + Maxim of Quantity \\
  $-->$ minimal number of entities introduced
\end{el}

We can see a direct application of the ``Minimality Assumption'' in
\ref{T8}:

\begin{el}
\ex {\it I was driving on the right hand side of the road when a vehicle
arriving in front of me in the curve was completely thrown off course.
Keeping as close as possible to the right, I wasn't able to avoid the car
which was coming with great speed.\/}  (T8)
\label{T8}
\end{el}

The text in \ref{T8} mentions only two vehicles.  The first
one, {\it W\/}'s, is implicit in {\it I was driving\/}.  The other one
is mentioned in two different expressions, {\it a vehicle arriving in
front of me in the curve\/} and {\it the car which was coming with
great speed\/}.  It is clear that the second expression is anaphoric
to the first, but this coreference is not explicit in the text.

The coreference is allowed first by the use of two compatible terms:
indeed a car is a particular type of vehicle.  This fact can be
extracted from a hierarchy of concepts which is part of the background
knowledge {\it K\/} (a {\it car\/} is the most typical kind of {\it
vehicle\/}).  Secondly, this coreference is licensed by the use of the
definite article, which allows the inference that the entity has
already been mentioned.  Finally, it is confirmed by the ``Minimality
Assumption'', which prevents the introduction of a third vehicle which
would not play any role in the scene.\footnote{
See \cite{Estival/Gayral/to appear} for a more detailed presentation and
more examples of this type of inferences permitted in our texts.}

\subsection{Parameter C}

In many of our texts, the accident is explicitly mentioned with verbs
such as {\it percuter, endommager, toucher, heurter\/} (``collide'',
``damage'', ``touch'', ``hit''), or with nouns such as {\it choc,
collision\/} (``impact'', ``collision'').  But this is not the case in
\ref{T15}, a text for which, if it was another type of narrative, we
might imagine other endings to the incident (e.g.\ {\it but I was able
to swerve and avoid it\/}).

\begin{el}
\ex {\it We were in Saint-Ouen, I was surprised by the person who
braked in front of me, not being able to change lanes, and the road
being wet, I couldn't stop completely in time\/} (T15)
\label{T15}
\end{el}

We can see here the effect of Parameter C:  since these texts are
accident reports, the series of events they relate must by default
contain an accident.  The interpretation of the texts often requires
the reconstruction of an impact between the two vehicles, otherwise
the incident which is described would not warrant the existence of the
report.

The existence of the impact can then be deduced from a combination of
several clues, some linguistic, some inferential.  Among the former,
we often find the combination of the negation with a verbal group of
the form ``can/be able to $+$ V'', for instance {\it I couldn't stop
completely in time\/} in \ref{T15}, or {\it I wasn't able to avoid the
car which was coming with great speed\/} in \ref{T8}.

\section{Argumentation}
\label{Argument}

So far, we have examined the texts from a purely factual point of
view.  Now, we take into account the argumentative aspect of these
texts.  Indeed, the authors know that these few lines, meant for their
insurance company, may contribute to the final decision about their
share of legal and financial liability.  So, they will try to
minimize their own responsibility.  There are two ways {\it W\/} can
justify his own behavior and make excuses for it:

\begin{itemize}
\item[A.] trying to push the blame onto his opponent by accusing him
of an abnormal behavior;

\item[B.] contrasting what was expected and what happened in reality,
by invoking unforeseable circumstances.
\end{itemize}

With either strategy, {\it W\/} must first show that he has done
everything that was required in the given circumstances and will
always try to appear as innocent as possible.

\subsubsection{Strategy A: Blaming the other driver}

With strategy A, {\it W\/} wants to suggest or to say explicitly that
the other driver is at fault.  Background knowledge {\it K\/} may be
used implicitly by {\it W\/} to suggest that his opponent has
misbehaved, and it also allows {\it R} to infer which behaviour is
``right'' and which one is ``wrong''.  For instance, in \ref{T11}, it
is clear, but not explicitly mentioned, that the driver of vehicle B
did something illegal, since in France one must pass on the left.

\begin{el}
\ex {\it The driver of vehicle B passed me on the right.\/}  (T11)
\label{T11}
\end{el}

In \ref{T12}, since vehicle B was not respecting the ground markings
it made a mistake in ``cutting back in'' on {\it W\/}'s vehicle.  Of
course, {\it W\/} has taken care to mention that his own vehicle was
in the correct lane.

\begin{el}
\ex {\it I was driving in my vehicle A in the right lane reserved for
vehicles going straight ahead.  Vehicle B was driving in the left lane
reserved for vehicles going left (ground markings with arrows).  It
cut back in on my vehicle.\/} (T12)
\label{T12}
\end{el}

These two examples show the importance of implicit knowledge, but
blaming the opponent can also be done explicitly.  Besides the example
in \ref{forcing}, where the truck driver is clearly said to be
responsible for the manoeuver, there are lexical clues, such as {\it
slalom\/} in \ref{slalom} (the driver of \ref{T11} is committing fault
after fault!), or {\it blinding\/} in \ref{aveugler}, suggesting that
the other driver was at fault.  In example \ref{right-of-way}, even
without the background knowledge of the French right-of-way rules, the
words {\it denies\/} and {\it right-of-way\/} strongly accuse the
other driver.  Driving at an excessive speed is of course a very
common characteristic of the other driver..., see \ref{aveugler} and
\ref{great-speed}.

\begin{el}
\ex
\begin{eli}
\exi {\it the latter turned left, \underline{forcing} me to steer left
to avoid it.\/} (T2)
\label{forcing}

\exi {\it According to the witness who was following me, the driver of
vehicle B was \underline{doing a slalom} between the cars.\/} (T11)
\label{slalom}

\exi {\it A vehicle with \underline{full white headlights}
\underline{blinding} us struck us \underline{with great speed} in the
back of the vehicle, taking us into a series of barrel rolls before
the vehicle stopped in a ditch.\/} (T10)
\label{aveugler}

\exi {\it Vehicle B coming from my left, I find myself at the
intersection, at moderate speed, about 40 km/h, when vehicle B hits my
vehicle, and \underline{denies} me the \underline{right-of-way} from
the right.\/} (T4)
\label{right-of-way}

\exi {\it at that moment vehicle B passed me \underline{with great
speed}\/} (T9)
\label{great-speed}
\end{eli}
\end{el}

\subsection{Strategy B: Blaming unforeseable circumstances}

Here, the indications are mostly at the lexical level (e.g.\ {\it
\^{e}tre surpris\/} ``to be surprised'') and make frequent use of the
negation.

\begin{el}
\ex
\begin{eli}
\exi {\it I was \underline{surprised} by the person who braked in
front of me, \underline{not being able to} change lanes, and the
\underline{road being wet}\/} (T15)

\exi {\it I \underline{didn't expect} that a driver would wish to pass
me for there weren't two lanes marked on the portion of the road where
I was stopped.}  (T5)
\end{eli}
\end{el}

The use of negation is also a favorite clue to indicate an opposition
between what should have happened and what actually occurred.

\begin{el}
\ex
\begin{eli}
\exi {\it I \underline{wasn't able} to avoid the car which was coming
with great speed.\/} (T8)
\label{speed}

\exi{\it and the road being wet, \underline{I wasn't able} to stop
completely in time.} (T15)
\label{wet}
\end{eli}

\label{negation}
\end{el}

Another device is the reverse of the metonymy conflating the vehicle
and its driver which we saw in \ref{param-A}.  For instance, in
\ref{T4}, it is not {\it W\/}, but the car which is the subject of the
two verbs, as if it was responsible for the events.

\begin{el}
\ex {\it on impact, and because of the slippery pavement, my vehicle
\underline{skids}, and \underline{hits} the metal railing around a
tree, whence a second front impact.\/} (T4)
\label{T4}
\end{el}

Of course the two strategies are not mutually exclusive; \ref{T15} and
\ref{T14} are actually instances of a mixture of both, in particular
\ref{T14} where {\it W\/} piles up all sorts of attenuating
circumstances and also emphasizes ({\it immediately put the brakes
on\/}) his own appropriate reactions.

\begin{el}
\ex
{\em I was driving at about 45 km/h in a \underline{small one-way}
street where cars were \underline{parked on both sides}.  Popping
\underline{suddenly} on my right coming out of a private building
garage, Mrs.Glorieux's vehicle was at a \underline{very short
distance} from my vehicle; passage being \underline{impossible}:
\underline{surprised}, I immediately put the brakes on but the impact
was unavoidable.}  (T14)
\label{T14}
\end{el}

Morover, the authors may choose to describe only that part of reality
which is in their favor, and the reader must thus be able to reconstruct
the items that were left out (intentionally or not).

With strategy B, linguistic clues include the use of reflexive verbs
({\it la porte s'est ouverte\/} ``the door opened'') and of the
passive voice ({\it j'ai \'{e}t\'{e} d\'{e}port\'{e}\/} ``I was thrown
off course'') instead of a plain active voice.  These constructions,
by suppressing the agent, suggest that {\it W\/} was not involved in
the course of events and cannot be held responsible for what happened.
We also find here all the adverbials and modifiers denoting unexpected
events or unusual states of affairs.  In addition to \ref{T14}, some
more examples are given in \ref{surprise}.

\begin{el}
\ex
\begin{eli}
\exi {\it but I hit the second car which hadn't \underline{yet} gone
through the stop-sign.\/} (T1)

\exi {\it I wasn't able to avoid the car which was coming with great
speed\/} (T8)

\exi {\it at that moment vehicle B passed me with great speed\/} (T9)
\label{T9}

\end{eli}
\label{surprise}
\end{el}

\subsection{Resolving Ambiguity and Drawing Inferences}
\label{ambiguity}

These texts provide a number of examples of clearcut ambiguity between
two situations A and B, which an argumentative type of justification
helps resolve.  The question that allows resolving the ambiguity is:
``What advantage would there be for {\it W\/} in implying situation A?
or in implying situation B?''.  We go in more details into some
examples.

\subsubsection{Lexical Ambiguity}

As shown in \ref{T8.rep}, the original French text of the example
given in \ref{T8} presents a common kind of ambiguity, since in
French, the word {\it droite\/} is ambiguous between the two
interpretations {\it right\/} and {\it straight\/}.\footnote{
If the adjective {\it droite\/} means {\it straight}, its opposite is
then {\it courbe/curved\/}, if it means {\it right\/}, the opposite is
then {\em gauche/left\/}.}

\begin{el}
\ex
Je roulais sur la \underline{partie droite} de la chauss\'{e}e (T8)

{\it I was driving on the (right-hand side)/(straight portion) of the road}
\label{T8.rep}
\end{el}

Here, even though the whole text can also be interpreted with the {\it
droite/straight\/} meaning, the {\it droite/right\/} interpretation is
more plausible.  However, only an argumentative type of reasoning can
lead {\it R\/} to prefer the latter.

Since the fact that in France one drives on the right is well-known,
in specifying that he was driving on the {\it right\/} side of the
road, {\it W\/} violates the Maxim of Quantity (i.e.\ not to say
anything superfluous) and therefore must be taken as intending to
convey some other information.  In this case, it must be in order to
assert that his behavior was conforming to the ``Rules of the Road'',
which is a pertinent fact to mention.  Here, informational redundancy
by itself carries some information which allows inference.

We can thus formulate the following rule:

\begin{el}
\ex
In case of ambiguity, prefer the interpretation which allows {\it R\/}
to infer a ``correct behavior'' on {\it W\/}'s part.
\label{ambiguity-rule}
\end{el}

\subsubsection{Time Reference Ambiguity}

In example \ref{T7}, repeated here as \ref{T7.rep} for convenience,
the use of the pluperfect {\it had switched on} is ambiguous.

\begin{el}

\ex {\it Being momentarily stopped in the right lane on Boulevard des
Italiens, \underline{I had switched my blinker on}; I was at a stop
and getting ready to change lanes.  Vehicle B coming from my left
squeezed too close to me and damaged the whole left front side.}  (T7)
\label{T7.rep}
\end{el}

The pluperfect implies that the process being talked about is
perceived with another past event as a point of reference, which may
not yet have been mentioned (the situation is exactly parallel in both
French and English).  Here, two different referential situations can
be envisaged, with two different consequences:

\begin{itemize}

\item If the accident itself is chosen as the point of reference,
switching the blinker on signals a future change of lanes.  It must
therefore be the left blinker.  This conclusion requires geometrical
reasoning:  ``If X is stopped in the right lane and if X wants to
change lanes, X can only go left''.

\item If the time of stopping is chosen as the point of reference,
switching the blinker on is prior to the time of stopping and thus
signals it.  It must then be the right blinker, since the vehicle is
in the right lane.
\end{itemize}

To make a decision, arguments of the ``Maxims'' type must be used.
{\it R\/} cannot assume that too much information is present in the
text.  The fact that the blinker would be switched on before stopping
would not be relevant since the accident occurred after the act of
stopping, when {\it W\/} started again.  On the other hand, the fact
that {\it W\/} did switch the blinker on before starting again is very
relevant from an argumentative point of view, since the message
conveyed is then ``{\it W\/} behaved in the right way and did what was
required''.\footnote{
The semantic interpretation of the verb itself can also help in making
the choice, see next section.}
Therefore, by appealing to the rule proposed in \ref{ambiguity-rule},
the first interpretation is chosen and {\it R\/} may conclude that
{\it W\/} had his left blinker on.

\subsubsection{Action or Intention?}
\label{intention}

Sometimes, the problem for {\it R\/} is to determine whether an action
presented as an intended future event has remained at a purely
intentional level or whether actions have been taken to attain it.
For instance, when the intended action belongs to a script with
sequential steps, the question arises whether some of the preparatory
actions belonging to the script have already been accomplished.

In \ref{T7.rep}, we saw that there were two possible choices for a
point of reference in the interpretation of the pluperfect.  In
addition, the verb {\it s'appr\^{e}ter \`{a}\/} can have several
interpretations.  Like {\it to get ready\/} (which we give here as its
translation), it can mean {\it to be about to\/} and then it is a
simple aspectual auxiliary focussing on the beginning of the action
(inchoative).  It can also have a more agentive interpretation and
then it means {\it to actively prepare for\/}.

In the inchoative {\it to be about to\/} interpretation, the action of
``switching the blinker on'' is an event independent of ``changing
lanes''; in the agentive {\it to prepare for\/} interpretation, that
same action corresponds to one of the preparatory acts.  But more
crucially, in the agentive interpretation, {\it W\/} may already have
started changing lanes and then probably would be at fault, while in
the inchoative reading, {\it W\/} would still be stopped and would be
innocent.

It seems that in most cases, such an intended future event is more
than simply intentional and that {\it W\/} has indeed already started
to act.  Otherwise it would not be possible to explain the accident in
\ref{T7.rep}, since there would be no reason for {\it W\/}'s car to
have been damaged if {\it W\/} hadn't already started turning left.

Similarly in the case of the texts given in \ref{T2} and \ref{T5}
below, the only plausible reconstruction of the accident requires
vehicle ``A'' to have already started the action which is presented as
an intention ({\em Wanting to pass a hauler\/} in \ref{T2} and {\it I
wanted to enter the second lane\/} in \ref{T5}).

\begin{el}
\ex {\it \underline{Wanting to pass a hauler} with its right blinker
on, the latter turned left, forcing me to steer left to avoid it.  The
car skidded on the wet pavement and struck a sidewalk then a fence
straight ahead.  The truck driver had indeed switched on his left
blinker, but the trailer was inverting the signal to the right.  Not
having touched me, the driver declared himself unconcerned by the
situation and refused to draw a report.  Having left my car to call a
mechanic, I came back to find it with the right back door bashed in
with no note left by the guilty party.}  (T2)
\label{T2}
\end{el}

\begin{el}
\ex {\it I was stopped at the intersection wishing to take the road on
which the intense traffic is going one-way in two lanes; as the last
vehicle of the flow was coming, \underline{I wanted to enter}
\underline{the second lane}, leaving the first one free for it.  The
moment I started, I heard the shock in the back; I wasn't expecting a
driver would wish to pass me for there weren't two lanes marked on the
portion of the road where I was stopped.}  (T5)
\label{T5}
\end{el}

Instead of using an imperfective verbal form (i.e.\ {\it \'{e}tant en
train de d\'{e}passer un semi-remorque\/} (``while passing a hauler'')
in \ref{T2}, or {\it j'\'{e}tais en train de tourner \`{a} gauche\/}
(``I was turning left'') in \ref{T5}) which would clearly indicate
that the action had already started, {\it W\/} chooses the intentional
form and in doing so, creates an ambiguity for {\it R\/}:  ``Had {\it
W\/} actually already done something or not?''.  This lack of
precision (or downright lie?)  is intentional and allows {\it W\/} to
try to lessen his responsibility, which will succeed if {\it R\/} opts
for a purely intentional reading of the verbal form.

\section{Conclusion}

In this paper we have tried to show the importance of situational,
cultural and textual presuppositions from the point of view of both
the writer {\it W\/} and the reader {\it R\/}.  As this work
constitutes a first step in the study of natural language semantics in
the context of an NLP project, the approach adopted here is an attempt
to automate the process of understanding these texts and deriving
inferences from them.  Crucial issues for NLP are how to define and
describe the different types of knowledge involved in the processes of
writing and reading texts, and how to establish rules that mimic the
reasoning involved in these activities.

Here, we take advantage of the specificity of the texts -- the authors
narrate events leading to a car accident while trying to lessen their
responsability -- to circumscribe the type of knowledge required and
to give some rules of interpretation, valid for this type of text, in
this type of context.  We have determined four parameters, and two
types of knowledge necessary for both the production and the
interpretation of these reports.  Two of these parameters (A and C)
and K belong to the factual domain, while the other two parameters (B
and D) and LC pertain to discourse.

For clarity of exposition, we have distinguished these two types of
characteristics in our texts by examining first the factual content of
the texts and then their argumentative aspect, but it is not always
easy to separate them and we can also ask whether there actually can
be a purely factual reading of a text that would not take into account
discourse and argumentation phenomena.

In any case, even if such a reading existed, it would be insufficient
to account for the inferences that the reader can and must make from
the textual data in order to reconstruct the events described by the
text and to determine each participant's role in it.  We have shown
for instance that inferences based on argumentation could often help
the reader clarify the text or choose between several interpretations.
We find here the well-known difficulty of precisely defining the
border between semantics and pragmatics.

It would be interesting to analyze the two corresponding texts by the
two opponents reporting the same accident in order to establish which
part of the information is objectively factual and shared by both
texts, and which part of the information is argumentatively biased,
thus better distinguishing the subjective part of both discourses.
The omission of information, which was mentioned as one of the
argumentative devices on the part of {\it W\/} and as a basis for
inference on the part of {\it R\/}, would then become an even more
important factor in the analysis.  Very few such pairs of texts are
available, but in the continuation of this project, we may try to do
some further work based on these.

Lastly, we have shown that some inferences rely on assessing the
relevance or the quality/quantity of the information given.  This
assessment itself refers to a norm which is shared by the community of
speakers and thus belongs to LC.  However, it remains extremely
difficult to define this norm in advance and this type of inference,
though crucial for language understanding, still appears beyond what
is currently possible in NLP.


\newpage
\pagestyle{empty}
\section*{Annex: Texts}
\subsection*{Text 1}

Me rendant \`{a} Beaumont sur Oise depuis Cergy.  Je me suis
retrouv\'{e}e \`{a} un carrefour juste apr\`{e}s la sortie Beaumont
sur Oise.  J'\'{e}tais \`{a} un stop avec 2 voitures devant moi
tournant \`{a} droite vers Mours.  Alors que la premi\`{e}re voiture
passait ce stop je fis mon contr\^{o}le \`{a} gauche et je
d\'{e}marrais mais je percutais la deuxi\`{e}me voiture qui n'avait
pas encore pass\'{e} le stop.

\subsection*{Text 2}

Voulant d\'{e}passer un semi-remorque qui tournait \`{a} droite, ce
dernier tourna \`{a} gauche m'obligeant \`{a} braquer \`{a} gauche
pour l'\'{e}viter.  La voiture a d\'{e}rap\'{e} sur la chauss\'{e}e
mouill\'{e}e et a percut\'{e} un trottoir puis un mur de cl\^{o}ture
en face.  Le conducteur du camion avait bien mis son clignotant \`{a}
gauche mais sa remorque inversait le signal sur la droite.  Ne m'ayant
pas touch\'{e} le conducteur s'est d\'{e}clar\'{e} hors de cause et
n'a pas voulu \'{e}tablir de constat.  Ayant quitt\'{e} ma voiture
pour appeler un d\'{e}panneur j'ai retrouv\'{e} celle-ci avec la
porti\`{e}re arri\`{e}re droite enfonc\'{e}e sans coordonn\'{e}es du
responsable.

\subsection*{Text 3}

Fort trafic \`{a} 17h15 Bd S\'{e}bastopol.  Je roulais entre deux
files de voitures arr\^{e}t\'{e}es quand l'une des voitures \`{a} ma
gauche a ouvert sa porte avant droite.  Pour l'\'{e}viter, j'ai fait
un \'{e}cart qui m'a fait toucher le v\'{e}hicule B avec l'arri\`{e}re
de ma moto ce qui a provoqu\'{e} ma chute.  Vu l'importance du trafic
\`{a} cette heure l\`{a} nous avons juste \'{e}chang\'{e} nos
assurances et noms ce qui explique que mon constat amiable ne soit
sign\'{e} que par moi.

\subsection*{Text 4}

V\'{e}hicule B venant de ma gauche, je me trouve dans le carrefour,
\`{a} faible vitesse environ 40 km/h, quand le v\'{e}hicule B, percute
mon v\'{e}hicule, et me refuse la priorit\'{e} \`{a} droite.  Le
premier choc atteint mon aile arri\`{e}re gauche, sous le choc, et
\`{a} cause de la chauss\'{e}e glissante, mon v\'{e}hicule d\'{e}rape,
et percute la protection m\'{e}tallique d'un arbre, d'o\`{u} un second
choc frontal.

\subsection*{Text 5}

J'\'{e}tais arr\^{e}t\'{e} \`{a} l'intersection d\'{e}sirant emprunter
la route o\`{u} la circulation intense s'effectue \`{a} sens unique
sur deux voies; lorsque le dernier v\'{e}hicule du flot arrivait, j'ai
voulu m'engager sur la deuxi\`{e}me file, lui laissant libre la
premi\`{e}re.  Au moment o\`{u} je d\'{e}marrais, j'ai entendu le choc
arri\`{e}re; je ne m'attendais pas \`{a} ce qu'un usager d\'{e}sire me
d\'{e}passer car il n'y avait pas deux voies mat\'{e}rialis\'{e}es sur
la portion de route o\`{u} je me trouvais \`{a} l'arr\^{e}t.

\subsection*{Text 6}

Mr C.Delon, abordant le carrefour, laissait le passage aux
v\'{e}hicules roulant sur la voie abord\'{e}e, car d'ordinaire se
trouve un feu \`{a} ce carrefour.  (hors fonctionnement ce
jour-l\`{a}).  Venant de derri\`{e}re moi, roulant dans le m\^{e}me
sens, dans la m\^{e}me file, Mr Oms n'a pas vu que j'\'{e}tais
arr\^{e}t\'{e} et a percut\'{e} fortement mon v\'{e}hicule, l'abimant
gravement.  De ce fait, j'ai subi (C.Delon) "le coup du lapin"; le
si\`{e}ge conducteur a \'{e}t\'{e} endommag\'{e}; les gendarmes se
sont rendus sur place; j'ignore s'ils ont \'{e}tabli un rapport.

\subsection*{Text 7}

Etant arr\^{e}t\'{e} momentan\'{e}ment sur la file de droite du
Boulevard des Italiens j'avais mis mon clignotant j'\'{e}tais \`{a}
l'arr\^{e}t et m'appr\^{e}tant \`{a} changer de file.  Le v\'{e}hicule
B arrivant sur ma gauche m'a serr\'{e} de trop pr\`{e}s et m'a
abim\'{e} tout le c\^{o}t\'{e} avant gauche.

\subsection*{Text 8}

Je roulais sur la partie droite de la chauss\'{e}e quand un
v\'{e}hicule arrivant en face dans le virage a \'{e}t\'{e}
compl\`{e}tement d\'{e}port\'{e}.  Serrant \`{a} droite au maximum, je
n'ai pu \'{e}viter la voiture qui arrivait \`{a} grande vitesse.

\subsection*{Text 9}

Nous roulions en ville sur une portion de route \`{a} deux voies
o\`{u} la vitesse est limit\'{e}e \`{a} 45km/h.  Je clignotais et
m'appr\^{e}tais \`{a} tourner \`{a} gauche vers le chemin de Condos.
A ce moment, le v\'{e}hicule B a doubl\'{e} \`{a} grande vitesse notre
v\'{e}hicule et s'est immobilis\'{e} sur le trottoir gauche de la
chauss\'{e}e apr\`{e}s m'avoir touch\'{e}.

\subsection*{Text 10}

Nous roulions sur une route \`{a} 90km/h.  Un v\'{e}hicule plein
phares blancs nous aveuglant nous a percut\'{e}s \`{a} grande vitesse
\`{a} l'arri\`{e}re du v\'{e}hicule, nous entra\^{\i}nant dans une
s\'{e}rie de tonneaux avant l'immobilisation du v\'{e}hicule dans un
foss\'{e}.

\subsection*{Text 11}

Le conducteur du v\'{e}hicule B me doublant par la droite a
accroch\'{e} mon pare-choc avant droit et m'a entra\^{\i}n\'{e} vers
le mur amovible du pont de Genevilliers que j'ai percut\'{e}
violemment.  D'apr\`{e}s les dires du t\'{e}moin qui me suivait, le
conducteur du v\'{e}hicule B slalomait entre les voitures.  Apr\`{e}s
m'avoir heurt\'{e}, il a pris la fuite et n'a pu \^{e}tre rejoint par
le t\'{e}moin cit\'{e}.

\subsection*{Text 12}

Je circulais \`{a} bors de mon v\'{e}hicule A sur la file de droite
r\'{e}serv\'{e}e aux v\'{e}hicules allant tout droit.  Le v\'{e}hicule
B circulait sur la voie de gauche r\'{e}serv\'{e}e aux v\'{e}hicules
allant \`{a} gauche (marquage au sol par des fl\`{e}ches).  Celui-ci
s'est rabattu sur mon v\'{e}hicule A me heurtant \`{a} l'arri\`{e}re
gauche.

\subsection*{Text 13}

Je roulais dans la rue Pasteur quand une voiture surgit de ma droite;
pour l'\'{e}viter, je me rabattais \`{a} gauche et freinais.  Je pus
l'\'{e}viter et mon r\'{e}troviseur heurte le sien.  La voiture
continue car elle n'eut rien et moi, je heurtai une benne qui
stationnait sur le c\^{o}t\'{e} de la chauss\'{e}e.  La benne n'a pas
\'{e}t\'{e} du tout endommag\'{e}e.  Ma voiture a \'{e}t\'{e}
touch\'{e}e \`{a} l'avant ainsi qu'au r\'{e}troviseur.

\subsection*{Text 14}

Je circulais \`{a} environ 45 km/h dans une petite rue \`{a} sens
unique o\`{u} stationnaient des voitures de chaque c\^{o}t\'{e}.
Surgissant brusquement sur ma droite sortant d'un parking d'immeuble,
le v\'{e}hicule de Mme Glorieux \'{e}tait \`{a} tr\`{e}s peu de
distance de mon v\'{e}hicule; le passage \'{e}tant impossible:
surpris, je freinais imm\'{e}diatement mais le choc fut
in\'{e}vitable.

\subsection*{Text 15}

Nous \'{e}tions \`{a} Saint-Ouen, j'ai \'{e}t\'{e} surprise par la
personne qui a frein\'{e} devant moi, n'ayant pas la possibilit\'{e}
de changer de voie et la route \'{e}tant mouill\'{e}e, je n'ai pu
m'arr\^{e}ter compl\`{e}tement \`{a} temps.

\subsection*{Text 16}

Je m'engageais (v\'{e}hicule A) dans une file de station service.  La
pompe \'{e}tant en panne, je reculais pour repartir lorsque j'ai
heurt\'{e} le v\'{e}hicule B qui s'\'{e}tait engag\'{e} \'{e}galement
dans la m\^{e}me file pour prendre de l'essence.

\subsection*{Text 17}

La conductrice de l'autre v\'{e}hicule et moi amor\c{c}ions le virage
sur la gauche dans un carrefour.  Nous \'{e}tions \`{a} la m\^{e}me
hauteur.  Nous nous sommes certainement rapproch\'{e}es et par
cons\'{e}quent percut\'{e}es, sa voiture s'embo\^{\i}tant dans la
mienne, son aile gauche dans l'avant lat\'{e}ral droit de ma voiture.

\end{document}